\documentclass{article}

\PassOptionsToPackage{comma,compress,round}{natbib}
\usepackage[preprint]{neurips_2025}

\usepackage[utf8]{inputenc} 
\usepackage[T1]{fontenc}    
\usepackage{nicefrac}       
\usepackage{microtype}      
\usepackage{xcolor}         
\usepackage{graphicx,verbatim}
\usepackage{amsmath}
\usepackage{amssymb}
\usepackage{arydshln}
\usepackage{bm}
\usepackage{booktabs,makecell,multirow,pifont,color, colortbl}
\usepackage{array}
\usepackage{tabularx} 

\usepackage[table]{xcolor} 
\usepackage{makecell}
\definecolor{lightmaroon}{RGB}{200,120,120}
\usepackage{makecell,xcolor}

\newcommand{\best}[1]{\cellcolor{blue!10}{#1}}
\newcommand{\up}[1]{{\scriptsize\textcolor{green!50!black}{($\uparrow$#1\%)}}}
\newcommand{\down}[1]{{\scriptsize\textcolor{red!70!black}{($\downarrow$#1\%)}}}
\newcommand{\neutral}{{\scriptsize\textcolor{gray}{(--)} }}
\usepackage{hyperref}
\usepackage{url}
\usepackage{float}

\title{Towards Application Aligned Synthetic Surgical Image Synthesis}

\author{%
  Danush Kumar Venkatesh$^{1,2}$ \quad Stefanie Speidel$^{2,3}$\\
  $^{1}$Department of Translational Surgical Oncology, NCT/UCC Dresden, a partnership between DKFZ, \\
  Faculty of Medicine and University Hospital Carl Gustav Carus, TUD Dresden,HZDR, Germany \\
  $^{2}$Department of Translational Surgical Oncology, NCT/UCC Dresden, Faculty of Medicine and \\
  University Hospital Carl Gustav Carus, TUD Dresden Germany\\
  $^{3}$The Centre for Tactile Internet with Human-in-the-Loop (CeTI), TUD Dresden\\
  \texttt{danushkumar.venkatesh@nct-dresden.de}
}

\begin{document}

\maketitle

\begin{abstract}
  The scarcity of annotated surgical data poses a significant challenge for developing deep learning systems in computer-assisted interventions. While diffusion models can synthesize realistic images, they often suffer from data memorization, resulting in inconsistent or non-diverse samples that may fail to improve, or even harm, downstream performance. We introduce \emph{Surgical Application-Aligned Diffusion} (SAADi), a new framework that aligns diffusion models with samples preferred by downstream models. Our method constructs pairs of \emph{preferred} and \emph{non-preferred} synthetic images and employs lightweight fine-tuning of diffusion models to align the image generation process with downstream objectives explicitly. Experiments on three surgical datasets demonstrate consistent gains of $7$--$9\%$ in classification and $2$--$10\%$ in segmentation tasks, with the considerable improvements observed for underrepresented classes. Iterative refinement of synthetic samples further boosts performance by $4$--$10\%$. Unlike baseline approaches, our method overcomes sample degradation and establishes task-aware alignment as a key principle for mitigating data scarcity and advancing surgical vision applications.
\end{abstract}

\section{Introduction}
Minimally invasive surgery (MIS) has gained increasing popularity in recent years due to its numerous benefits, including shorter recovery times, reduced postoperative pain, improved surgical dexterity and a lower risk of infection~\citep{dagnino2024robot}. The primary objective of MIS is to minimize the number and size of incisions; however, this also introduces challenges for surgeons. Procedures are typically performed by observing 2D endoscopic images on a monitor, which restricts the field of view and eliminates depth perception. These constraints highlight the need for computational methods that can function as assistive technologies, supporting surgeons during interventions. With the rapid advancements in deep learning (DL), there is an opportunity to develop such systems to improve surgical safety and efficiency. A key application area is surgical scene understanding, which involves tasks such as identifying anatomical structures, segmenting target tissues, or issuing warnings about critical structures (e.g., arteries) near surgical instruments. For instance, DL methods can help localize hidden tumor tissue by overlaying $3$D anatomical structures onto the intra-operative scene, thereby providing surgeons with crucial real-time guidance. Such visualizations have proven to improve surgical outcomes~\citep{wagner2012three}.

Despite their promise, current DL approaches in the surgical domain rely on supervised learning, which requires large and diverse annotated datasets. Acquiring such surgical datasets are particularly challenging due to patient privacy concerns, logistical constraints in the operating room, and strict data protection regulations across clinical centers~\citep{maier2017surgical}. This creates a paradox in the field: while large-scale datasets are essential for building robust DL models, access to such data remains scarce.

\begin{figure}
  \centering
  \includegraphics[height=3.5cm, width=0.95\textwidth]{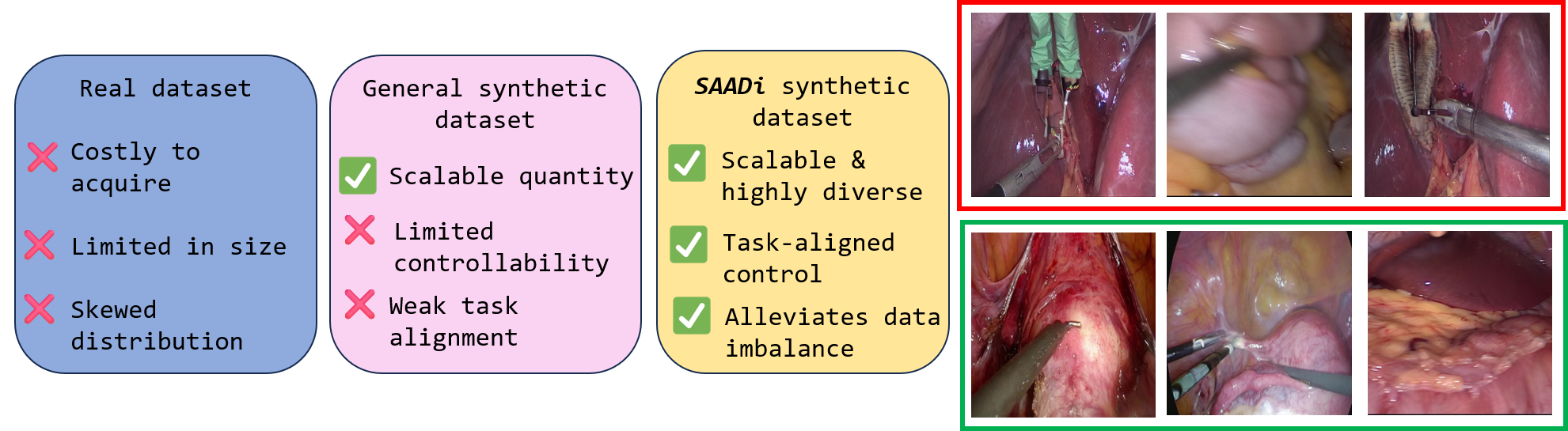}
  \caption{Comparison of real, general synthetic, and our (SAADi) datasets. Real datasets are often limited and imbalanced, and baseline generative models frequently yield inconsistent samples, such as blurred images (highlighted in the red box). In contrast, our method (SAADi) produces task-aligned, diverse, and realistic synthetic images that better capture anatomical structures (green box).}
  \label{fig:intro}
\end{figure}

To address the scarcity of real surgical data, increasing attention has been given to the use of synthetic datasets. Recent advances in generative modeling, particularly diffusion models (DMs)~\citep{ho2020denoising,dhariwal2021diffusion}, have shown remarkable ability to produce high-quality, photorealistic images. In the surgical domain, DMs have been trained on real-world datasets to generate clinically plausible synthetic images, which can then be combined with real data to enhance the performance of downstream deep learning (DL) models~\citep{nwoye2025surgical,frisch2023synthesising,venkatesh2025data}. However, the use of synthetic data poses several key challenges. First, controlling the composition of generated images is crucial, as this directly influences the effectiveness of downstream models. Second, due to the typically small size of surgical datasets, DMs are prone to overfitting, often replicating samples from the training set~\citep{somepalli2023understanding,chen2024towards} or producing undesirable configurations (see Fig.~\ref{fig:intro}). Incorporating such low-quality or redundant samples into training pipelines has been shown to provide limited benefit, and in some cases may even degrade performance~\citep{alaa2022faithful,azizi2023synthetic}, thereby undermining the purpose of synthetic data augmentation. These limitations raise critical questions: how can we effectively control DMs to ensure that the generated data contributes meaningfully to downstream tasks, and how can we guarantee that the resulting synthetic data is indeed beneficial? This motivates the central goal of our work: to develop a diffusion-based framework that generates synthetic surgical data that is directly beneficial for downstream models.  

To this end, we propose \emph{SAADi}, \textbf{S}urgical \textbf{A}pplication-\textbf{A}ligned \textbf{Di}ffusion, a framework for synthetic image generation that produces not only realistic samples but also data explicitly aligned with downstream task performance. To the best of our knowledge, this is the first work to introduce application-aligned diffusion for surgical image synthesis.
 We build on Stable Diffusion (SD)~\citep{rombach2022high}, a latent diffusion model, and introduce a framework in which the preferences of a downstream model explicitly guide image generation. Our approach is inspired by Diffusion-Direct Preference Optimization (DDPO)~\citep{wallace2024diffusion}, where DMs are fine-tuned on human preference data to improve aesthetic quality and prompt adherence. In contrast, we replace human supervision with automatically constructed preference pairs: we generate a large set of synthetic images using SD (trained on real surgical data), evaluate them with a downstream model (e.g., classification or detection), and retain or discard samples based on a predefined threshold. These preference pairs of \emph{preferred} and \emph{non-preferred} instances are constructed solely from the synthetic data and subsequently used to fine-tune SD with LoRA~\citep{hu2022lora}, introducing only minimal overhead. 
 Our approach directly addresses the dataset scarcity paradox in surgical science by ensuring that synthetic data is diverse, clinically relevant, and actively improves downstream surgical computer vision tasks.
We summarize our contributions as follows:
\begin{enumerate}
    \item We introduce \textbf{SAADi}, the first application-aligned diffusion framework for surgical image synthesis, explicitly designed to improve downstream performance.  
    \item We introduce an innovative preference-pairing strategy that uses only synthetic images, enabling alignment of diffusion models through lightweight LoRA fine-tuning without additional human supervision.  
    \item We provide extensive experiments on three surgical datasets and two key downstream tasks (classification and segmentation), demonstrating consistent improvements, with performance gains of up to $15\%$, particularly in underrepresented classes.  
\end{enumerate}
\section{Related Work}
\subsection{Diffusion models} Diffusion models (DMs)~\citep{sohl2015deep} have revolutionized image synthesis with their superior image quality compared to generative adversarial networks (GANs)~\citep{goodfellow2014generative}. Particularly, Latent Diffusion Models (LDMs)~\citep{rombach2022high} extend this framework by performing the diffusion process in a compressed latent space, thereby significantly reducing computational costs while maintaining high image fidelity. Stable Diffusion (SD)~\citep{rombach2022high} is a large-scale implementation of LDMs trained on natural image datasets, where image generation is conditioned on text prompts. This conditioning is achieved by encoding text inputs into latent vectors using pre-trained language models such as CLIP~\citep{radford2021learning}. Owing to its strong generative capabilities and open-source availability, SD has emerged as one of the most widely adopted LDM variants. In this work, we build our framework on top of the SD model.
\subsection{Synthetic surgical images} Laparoscopic image synthesis has been focused predominantly on image-to-image (I2I) translation methods. For example, computer-simulated surgical images, phantom data, and segmentation maps have been employed with GANs to synthesize realistic surgical images and videos~\citep{chen2019learning,sankaranarayanan2018learning,pfeiffer2019generating,Rivoir_2021_ICCV,venkatesh2024exploring,yoon2022surgical,sharan2021mutually,marzullo2021towards}. More recently, Stable Diffusion (SD)-based I2I methods were explored in~\citep{kaleta2024minimal,venkatesh2024cd,martyniak2025simuscope,venkatesh2025data}. Importantly, large quantities of synthetic images are generated, although their quality can sometimes be detrimental for the downstream task~\citep{venkatesh2024exploring,frisch2023synthesising}. Beyond surgical applications, diffusion models have rapidly gained traction for medical image generation, particularly in MRI and CT~\citep{dorjsembe2022three,khader2022medical,lyu2022conversion}. However, these images differ broadly in modality from the surgical images.
\subsection{Controllable generation}
Effective control of diffusion models (DMs) is critical for customizing generated images. Text-based editing has been explored through prompt engineering and manipulation of CLIP features~\citep{avrahami2022blended,brooks2023instructpix2pix,gafni2022make,hertz2022prompt,kawar2023imagic}, but such approaches are less suitable in the surgical domain where detailed textual descriptions are scarce. In contrast, spatial control can be achieved with conditional images processed via adapter networks resembling the U-Net backbone in latent diffusion models (LDMs)~\citep{zhang2023adding}, or through lightweight adapters such as T2I-Adapter~\citep{mou2023t2iadapter}. While these methods provide strong controllability, they require substantial computational resources, large annotated datasets, and long training times for surgical adaptation. In this work, we take a simpler approach: we leverage existing datasets and implicitly guide the generation process by selecting task-relevant samples from the training distribution, thereby directly improving downstream performance.  
\section{Background}
\begin{figure}
  \centering
  \includegraphics[height=3.7cm, width=0.95\textwidth, keepaspectratio]{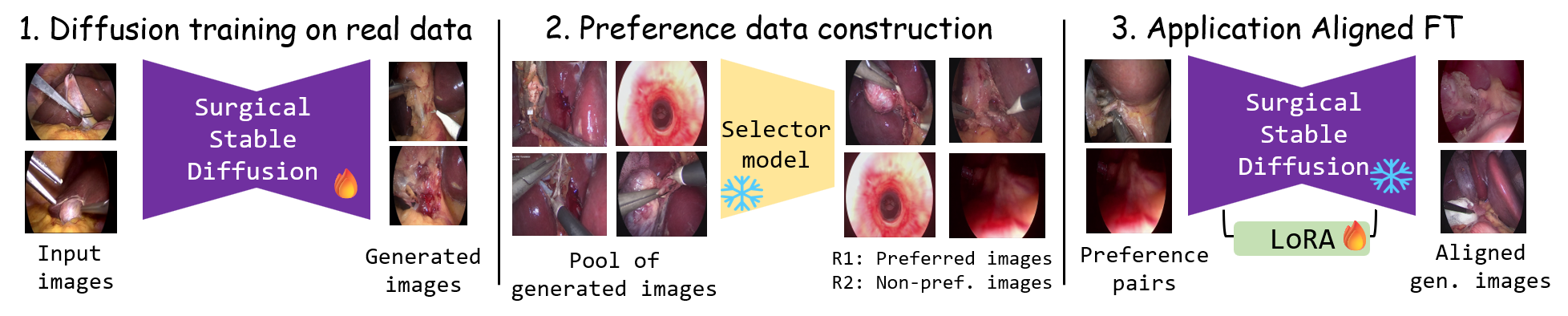}
  \caption{\textbf{Overview of the SAADi framework.} In Stage~$1$, we train a Surgical Stable Diffusion (SSD) model on real surgical images with text prompts and generate a large pool of synthetic images. In Stage~$2$, a selector model evaluates these synthetic images and separates them into \emph{preferred} and \emph{non-preferred} sets. In Stage~$3$, the resulting preference pairs are used to fine-tune the SSD model with LoRA adapters, aligning it with the preferences of the downstream task. The fine-tuned model is then used to sample task-aligned synthetic images.}
  \label{fig:model}
\end{figure}

In this section, we present our approach (SAADi) for generating synthetic data that is explicitly aligned with downstream tasks. The framework consists of two main stages. In the first stage, we train a Stable Diffusion (SD) model on real-world surgical datasets to learn the underlying data distribution. In the second stage, we generate synthetic samples from the trained model and construct preference pairs to guide alignment with the downstream task. To achieve this, we employ a selection model trained for the specific downstream task and run inference on the generated data. Based on a predefined threshold, each synthetic sample is categorized into a \emph{preferred} or \emph{non-preferred} set. Using this preference dataset, we fine-tune the SD model obtained from the first stage and subsequently sample from the refined model to obtain diverse and task-relevant synthetic images. An overview of the pipeline is shown in Fig.~\ref{fig:model}.

\subsection{Diffusion models}
Given samples from a data distribution $q(x_{0})$, and a noise scheduling function $\alpha_{t}$ and $\sigma_{t}$ (as defined in~\citep{ho2020denoising}), diffusion models are generative models $p_{\theta}(x_{0})$ trained to progressively denoise corrupted data. The training objective is defined as  
\begin{equation}
    \mathcal{L}_{\mathrm{DM}} = \mathbb{E}_{x_{0}, \epsilon, t, x_{t}} 
    \left[ \, \lVert \epsilon - \epsilon_{\theta}(x_{t}, t, P) \rVert^{2}_{2} \, \right],
\end{equation}
where $\epsilon \sim \mathcal{N}(0, \mathbf{I})$, $t \sim \mathcal{U}(0, T)$, and $x_{t} \sim q(x_{t}\mid x_{0}) = \mathcal{N}(x_{t}; \alpha_{t} x_{0}, \sigma_{t}^{2}\mathbf{I})$. Here, $P$ denotes the text prompt. In the Stable Diffusion (SD) model~\citep{rombach2022high}, an encoder $E$ maps an input image $x_{0}$ into a latent space where the diffusion process is carried out, and a decoder $D$ reconstructs the denoised latent back into the pixel space. We call this model \emph{Surgical Stable Diffusion} (SSD) and sample images from it using $P$.

Similarly, we also employ the \emph{Surgical Stable Inpaint (SSI)} model from~\citet{venkatesh2025data}, which is trained for inpainting-based synthesis. Given an image $x_{0}$ and a mask $m$, the model is trained to synthesize realistic texture within the masked region. Formally, in the forward process, the masked input $\tilde{x}_{t}$ is constructed as  
$\tilde{x}_{t} = x_{t} \odot m + x_{0} \odot (1 - m)$,
where $x_{t}$ denotes the noised image at timestep $t$. The denoising network $\epsilon_{\theta}$ is trained with a modified objective:  
\begin{equation}
    \mathcal{L}_{\mathrm{SSI}} = \mathbb{E}_{\epsilon \sim \mathcal{N}(0, \mathbf{I}),\, t} 
    \left[ \, \lVert \epsilon - \epsilon_{\theta}(\tilde{x}_{t}, t, P, m) \rVert^{2}_{2} \, \right].
\end{equation}
Since the training objective is localized to the masked regions, SSI learns to generate organ-specific textures conditioned jointly on the segmentation mask and the text signal.
\subsection{Preference data creation}
Let $\mathcal{R} = \{ r_{i} \}_{i=1}^{N}$ denote the set of real surgical images, and $\mathcal{G}_{s} = \{ g_{j} \}_{j=1}^{M}$ the pool of synthetic images generated by the diffusion model (SSD or SSI). A downstream model $f(\cdot)$ is trained on $\mathcal{R}$ to map images to task-specific outputs (e.g., class labels or segmentation masks). To construct preference pairs, each synthetic sample $g_{j} \in \mathcal{G}_{s}$ is evaluated using $f$, producing a score $s_{j} = f(g_{j})$. Given a predefined threshold $h$, we separate $\mathcal{G}_{s}$ into preferred and non-preferred subsets:
\begin{equation}
    \mathcal{G}_{p} = \{ g_{j} \in \mathcal{G}_{s} \mid s_{j} \geq h \}, 
    \quad
    \mathcal{G}_{np} = \{ g_{j} \in \mathcal{G}_{s} \mid s_{j} < h \}.
\end{equation}
From these partitions, we define the preference dataset as
$
    \mathcal{D}_{\text{pref}} = \{ (g_{p}, g_{np}) \mid g_{p} \in \mathcal{G}_{p}, \; g_{np} \in \mathcal{G}_{np} \},
$
which is subsequently used to fine-tune the diffusion model for alignment.
\subsection{Application aligned optimization}
To generate synthetic data from the SSD or SSI models ($\epsilon_{\text{ref}}$) that is beneficial for downstream tasks, it is necessary to align these models with the preferred samples identified by the downstream model. Our approach, \textbf{SAADi} replaces human labels in DDPO with downstream task evaluations and construct preference pairs from synthetic data, as described in the previous step. The objective is to learn a new model $\epsilon_{p\theta}$ whose generations are explicitly aligned with these preferences by fine-tuning on pairs of \emph{preferred} and \emph{non-preferred} synthetic images.
Let $(x_0^{p}, x_0^{n}) \sim \mathcal{D}_{\text{pref}}$ be a preferred/non-preferred pair. For $t \sim \mathcal{U}(0,T)$, sample $\epsilon^{p}, \epsilon^{n} \sim \mathcal{N}(0,\mathbf{I})$ and the forward diffusion is defined via,
\[
x_t^{p} = \alpha_t x_0^{p} + \sigma_t \epsilon^{p}, 
\qquad
x_t^{n} = \alpha_t x_0^{n} + \sigma_t \epsilon^{n}.
\]
Let per-preference sample differences (relative to a fixed reference denoiser $\epsilon_{\mathrm{ref}}$) be defined as
\[
\Delta^{p}_{\theta}(t) \!=\! \|\epsilon^{p} - \epsilon_{p\theta}(x_t^{p}, t)\|_2^{2} - \|\epsilon^{p} - \epsilon_{\mathrm{ref}}(x_t^{p}, t)\|_2^{2},
\quad
\Delta^{n}_{\theta}(t) \!=\! \|\epsilon^{n} - \epsilon_{p\theta}(x_t^{n}, t)\|_2^{2} - \|\epsilon^{n} - \epsilon_{\mathrm{ref}}(x_t^{n}, t)\|_2^{2}.
\]
With the logistic function $\sigma(u)=1/(1+e^{-u})$, we define the loss as
\begin{equation}
\label{eq:pref-loss}
\mathcal{L}_{\text{SAADi}}(\theta)
= -\,\mathbb{E}_{(x_0^{p},x_0^{n}) \sim \mathcal{D}_{\text{pref}},\, t,\, \epsilon^{p},\, \epsilon^{n}}
\left[
\log \sigma\!\left(
-\,\beta\,\big(\Delta^{p}_{\theta}(t) - \Delta^{n}_{\theta}(t)\big)
\right)
\right],
\end{equation}
where $\beta$ is a weighting term. For additional details the readers can refer to~\citet{wallace2024diffusion}. Once trained, the model generates synthetic data that are explicitly aligned with downstream task preferences, ensuring their utility for the downstream task.
\section{Experiments}

In this section, we present our experimental setup, including the datasets and downstream models. We evaluate our approach on three surgical datasets across two tasks: (i) multi-class classification of anatomical structures and surgical tools, and (ii) binary segmentation of anatomical structures and tools. Our primary objective is to assess the utility of the generated synthetic data by measuring its impact on downstream performance. An overview of the datasets and tasks is provided in Tab.~\ref{tab:dataset_summary}.

\begin{table}[t]
\centering
\caption{\textbf{Overview of the datasets}. The different anatomies, tools and diffusion baselines used for evaluation is listed here.}
\label{tab:dataset_summary}
\setlength{\tabcolsep}{4pt}
\renewcommand{\arraystretch}{1.15}
\resizebox{\linewidth}{!}{%
\begin{tabular}{l c c c >{\raggedright\arraybackslash}p{4.2cm} >{\raggedright\arraybackslash}p{4.2cm} c}
\toprule
\textbf{Dataset} & \textbf{Procedure} & \textbf{Train/Test} & \textbf{\#Classes} & \textbf{Task: Anatomy} & \textbf{Task: Tools} & \textbf{Diffusion Baseline} \\
\midrule
LapGyn~\citep{lapgyn} & Gynecological surgery & 1014 / 438 & 5 organs {+} tools & Colon, Liver, Ovary, Oviduct, Uterus & Graspers, etc. & SSD (Surgical Stable Diffusion) \\
Endoscapes~\citep{murali2023endoscapes} & Cholecystectomy & 343 / 100 & 5 organs {+} tools & Cystic plate, Calot’s triangle, Cystic artery, Cystic duct, Gallbladder & Surgical tools & SSI (Surgical Stable Inpaint) \\
AutoLaparo~\citep{wang2022autolaparo} & Hysterectomy & 1100 / 500 & 1 organ {+} 4 tools & Uterus & Grasping forceps, Dissecting forceps, LigaSure, Electric hook & SSI (Surgical Stable Inpaint) \\
\bottomrule
\end{tabular}}
\end{table}

\paragraph{Evaluation scheme.} 
We design three evaluation settings to analyze the impact of synthetic datasets on downstream performance:  

\begin{enumerate}
    \item \textbf{Baseline comparison.} We generate synthetic data using the baseline models (SSD and SSI) and compare their performance against our approach (SAADi) across the downstream tasks described earlier. To ensure fairness, we add the same number of synthetic samples as real samples present in the training set. 

    \item \textbf{Data scaling.} We study the scaling behavior of synthetic data by adding multiples of the training set size ($2\times, 3\times, 4\times$) in synthetic samples from each method to the real dataset. This experiment assesses the impact of increasing synthetic data on downstream performance.  

    \item \textbf{Iterative refinement.} We further investigate whether synthetic data quality can be improved through refinement. In the first round, we train a downstream model on real data combined with synthetic images generated by SAADi. This trained model is then used as the selection model to re-score the initial pool of synthetic samples. Based on this updated scoring, we perform a second round of SAADi fine-tuning and generate a new set of images. The downstream models are subsequently evaluated on this refined dataset. This iterative process demonstrates that once useful synthetic data is introduced, downstream models improve, enabling the selection of stronger and more informative samples in subsequent rounds.
\end{enumerate}  

\paragraph{Surgical datasets} 
For the multi-class classification task, we use the LapGyn (LG) dataset~\citep{lapgyn}, which consists of laparoscopic gynecological procedures. The dataset includes five anatomical structures: colon, liver, ovary, oviduct, and uterus. The training set contains $1014$ images, while the test set contains $438$ images. Notably, the dataset is imbalanced across classes. The task is defined as classifying anatomical structures and surgical tools in a given surgical scene.  

For segmentation, we employ two datasets. First, the Endoscapes (Ed) dataset~\citep{murali2023endoscapes}, where we use only the segmentation split comprising $343$ training images and $100$ test images from laparoscopic cholecystectomy. This dataset allows us to evaluate performance in a low-data regime. The annotated structures include the cystic plate, Calot’s triangle, cystic artery, cystic duct, gallbladder, and surgical tools.  

Second, the Autolaparo (AL) dataset~\citep{wang2022autolaparo}, which contains data from $21$ patients undergoing laparoscopic hysterectomy, with $1100$ images for training and $500$ images for testing. The annotated classes include surgical tools such as grasping forceps, dissecting forceps, LigaSure, and electric hook, along with the uterus as the anatomical structure. Since patient diversity is a crucial factor in the surgical domain, we utilize a held-out, patient-specific test set to assess the models.  

These datasets were chosen to investigate the role of synthetic data under both class-imbalanced and resource-constrained conditions. 

\begin{figure}
  \centering
  \includegraphics[height=5cm,width=0.95\textwidth]{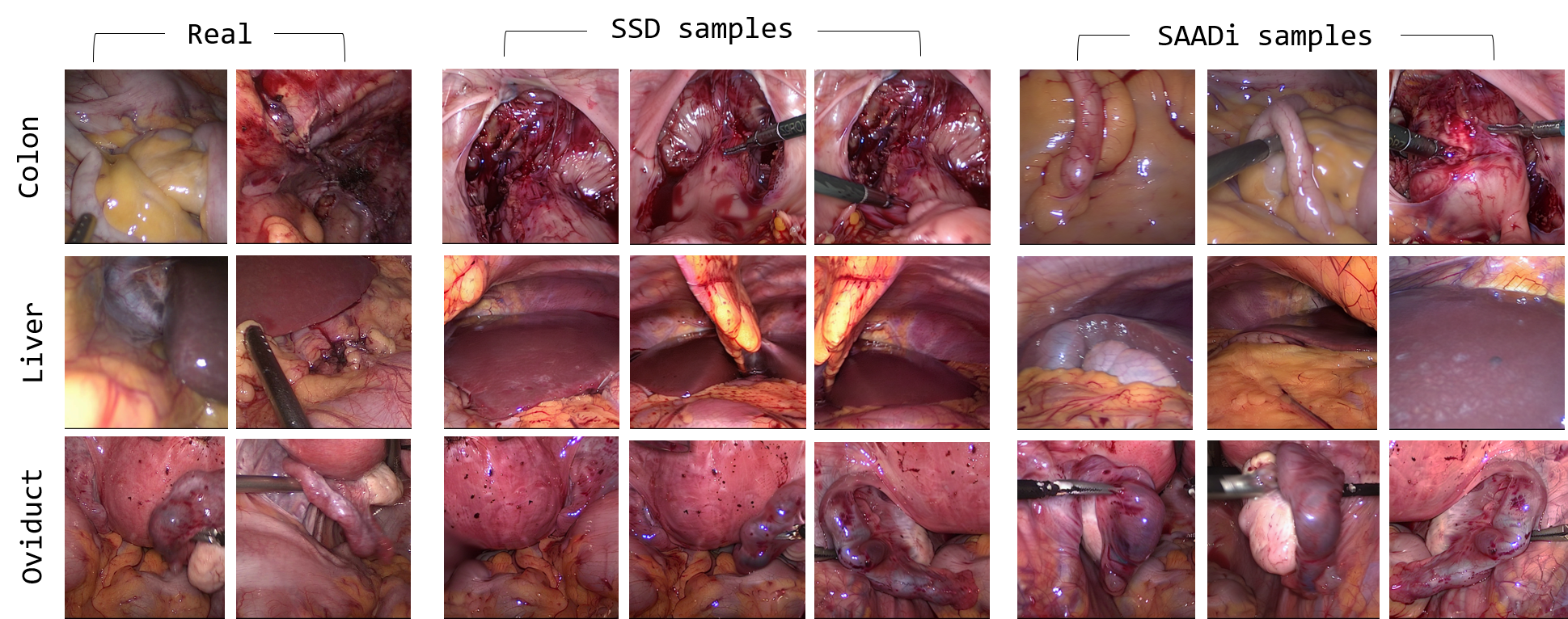}
  \caption{\textbf{Qualitative results} on the LapGyn dataset. Synthetic images from the baseline SSD model show limited diversity, with minimal variation across samples. In contrast, images generated by our method (SAADi) are diverse, application-aligned, and visually realistic, closely resembling real data.}
  \label{fig:auto_images}
\end{figure}

\paragraph{Baselines \& Models} 
As baselines for diffusion models, we trained the Surgical Stable Diffusion (SSD) model on the real surgical images from the LG dataset with the text prompts constructed as \emph{``An image of <organ/tool> in laparoscopic gynecological surgery''}. We sample images from this model and add them with real dataset as a baseline against \emph{SAADi}.

For the segmentation datasets, we used an inpainting model, \emph{Surgical Stable Inpaint} (SSI)~\citep{venkatesh2025data}, as the task involves generating organ and tool textures conditioned on masks. 
All diffusion models were trained using only the training splits of each dataset for $3000$ steps with AdamW~\citep{loshchilov2017decoupled}. For the inpainting models, we used $30$ denoising steps during the generation process. Subsequently, \emph{SAADi} fine-tuning was performed for $1500$ steps, requiring approximately $8$ minutes of training on a single $24$GB GPU. We change the prompts correspondingly for each dataset and organ or tool. 

For the downstream tasks, we adopted three architectures for classification: ResNet-$50$~\citep{he2016deep}, ConvNeXT-S~\citep{liu2022convnet}, and ViT-S~\citep{dosovitskiy2020image}; and three for segmentation: DeepLabV3 (DV3)~\citep{chen2017rethinking}, SegFormer~\citep{xie2021segformer}, and UPerNet~\citep{xiao2018unified}. To mitigate class imbalance, we applied pixel weighting and inverse-frequency balancing, combined with standard data augmentations, during training on real datasets. Employing a diverse set of architectures allowed us to reduce bias toward any single model. For evaluation, we report the F$1$ and Dice score for classification and segmentation tasks respectively.. 

\section{Results \& Discussion}
\begin{table}
\caption{\textbf{Classification of anatomies in the LapGyn dataset}. Reported values are F$1$ scores. Imbalanced classes are \colorbox{lightmaroon!20}{highlighted}, and the \best{best scores} are shown in \best{blue}. The addition of synthetic images from our approach (SAADi) improves performance by approximately $4$--$9\%$ compared to training on the real dataset alone.}
\label{tab:class_lap}
\begin{center}
{\small
\resizebox{\linewidth}{!}{
\begin{tabular}{l cc ccccc}
\toprule
Method & Training data & \cellcolor{lightmaroon!20}Colon & \cellcolor{lightmaroon!20}Liver & Ovary & \cellcolor{lightmaroon!20}Oviduct & Uterus & Mean\\
\midrule
\multirow{3}{*}{ResNet-$50$} 
  & Only Real  & $0.10$ & $0.22$ & $0.38$ & $0.06$ & $0.26$ & $0.20$ \\
  & Real $+$ SSD   & $0.10$ & \best{$0.25$} & $0.37$ & $0.09$ & $0.23$ &  $0.21$ \up{1}\\
  & Real $+$ SAADi    & \best{$0.15$} & $0.24$ & \best{$0.40$} & \best{$0.13$} & \best{$0.31$} & $0.24$ \up{4} \\
\midrule
\multirow{3}{*}{ConvNeXT/S} 
  & Only Real       & $0.10$ & $0.23$ & $0.30$ & $0.15$ & $0.42$ & $0.24$ \\
  & Real $+$ SSD   & $0.11$ & $0.23$ & $0.35$ & $0.16$ & $0.41$ & $0.24$ \neutral  \\
  & Real $+$ SAADi   & \best{$0.16$} & \best{$0.26$} & \best{$0.41$} & \best{$0.22$} & \best{$0.48$} & $0.31$ \up{7}\\
\midrule
\multirow{3}{*}{ViT/S} 
  & Only Real       & $0.10$ & $0.25$ & $0.32$ & $0.11$ & $0.37$ & $0.23$\\
  & Real $+$ SSD    & $0.14$ & $0.23$ & $0.34$ & $0.22$ & \best{$0.39$} & $0.26$ \up{3} \\
  & Real $+$ SAADi   & \best{$0.19$} & \best{$0.26$} & \best{$0.45$} & \best{$0.24$} & $0.38$ & $0.32$ \up{9} \\
\bottomrule
\end{tabular}
}}
\end{center}
\end{table}

\paragraph{Addition of synthetic samples} The qualitative results are shown in Fig.~\ref{fig:auto_images}. The results for the classification task on the LG dataset are presented in Tab.~\ref{tab:class_lap}. We observe that adding synthetic samples from the baseline model provides modest improvements in classification performance. In contrast, our approach, SAADi, yields substantial gains, with overall improvements of $7\%$ and $9\%$ for the ConvNeXT and ViT architectures, respectively. Notably, SAADi achieves significant improvements in the imbalanced classes, including an increase of more than $13\%$ for the oviduct class. Although the same number of samples is added across all baselines, the results demonstrate that SAADi is particularly effective in addressing class imbalance and improving performance on underrepresented categories.  

\begin{table}[!htbp]
\caption{\textbf{Anatomy segmentation in the Endoscapes dataset.} Synthetic samples generated by the baseline method (SSI) often introduce degenerate cases, leading to degraded performance. In contrast, images produced by our approach (SAADi) yield consistent improvements across all evaluated downstream models. Dice scores is reported.
}
\label{tab:endo}
\begin{center}
{\small
\resizebox{\linewidth}{!}{
\begin{tabular}{l cc ccccccc}
\toprule
Method & Training data & Cystic plate & Calot triangle & Cystic artery &Cystic duct& Gall bladder& Tool & Mean\\
\midrule
\multirow{3}{*}{DV3} 
  & Only Real       & $0.38$ & \best{$0.37$} & $0.42$ & $0.47$ & $0.70$ & $0.61$ & $0.49$ \\
  & Real $+$ SSI   & $0.36$ & $0.34$ & $0.43$ & $0.48$ & $0.64$ & \best{$0.66$} & $0.48$ \down{1}\\
  & Real $+$ SAADi    & \best{$0.41$} & $0.36$ & \best{$0.44$} & \best{$0.50$} & \best{$0.73$} & $0.61$ & $0.51$ \up{2} \\
\midrule
\multirow{3}{*}{Segformer} 
  & Only Real       & $0.40$ & $0.36$ & $0.30$ & $0.41$ & $0.68$ & $0.62$ & $0.46$ \\
  & Real $+$ SSI   & $0.39$ & $0.36$ & $0.38$ & $0.37$ & $0.64$ & $0.55$ & $0.44$ \down{2} \\
  & Real $+$ SAADi   & \best{$0.41$} & \best{$0.49$} & \best{$0.40$} & \best{$0.52$} & \best{$0.70$} & \best{$0.63$} & $0.52$ \up{6}\\
\midrule
\multirow{3}{*}{UPerNet} 
  & Only Real       & $0.33$ & $0.27$ & $0.37$ & $0.38$ & $0.54$ & $0.41$ & $0.38$\\
  & Real $+$ SSI    & \best{$0.41$} & $0.42$ & $0.32$ & $0.39$ & $0.52$ & $0.48$ & $0.42$ \up{4} \\
  & Real $+$ SAADi   & $0.40$ & \best{$0.44$} & \best{$0.46$} & \best{$0.40$} & \best{$0.57$} & \best{$0.61$} & $0.48$ \up{10} \\
\bottomrule
\end{tabular}
}}
\end{center}
\end{table}

Tab.~\ref{tab:endo} and Tab.~\ref{tab:auto} show the results of the segmentation models on the Ed and AL datasets. We observe that adding synthetic samples from the SSI baseline leads to a decline in performance for both SegFormer and DeepLabV3, suggesting that image generation with structure-specific constraints alone is insufficient to meet the requirements of downstream models. A modest improvement of $4\%$ is observed with UPerNet, further highlighting the importance of evaluating synthetic data across multiple architectures.  

\begin{table}
\caption{\textbf{Segmentation of tools in the Autolaparo dataset.} The SSI model fails to generate valid synthetic data, and its inclusion with the real dataset reduces performance, particularly for surgical tools. In contrast, our approach (SAADi) provides smaller yet consistent benefits when synthetic images are added.}
\label{tab:auto}
\begin{center}
{\small
\resizebox{\linewidth}{!}{
\begin{tabular}{l cc cccccc}
\toprule
Method & Training data & Grasping forceps & Liga Sure & Dissecting forceps & Electric hook & Uterus & Mean\\
\midrule
\multirow{3}{*}{DV3} 
  & Only Real       & $0.61$ & \best{$0.90$} & $0.74$ & $0.59$ & $0.73$  & $0.71$ \\
  & Real $+$ SSI   & $0.56$ & $0.48$ & $0.53$ & $0.54$  & $0.74$ & $0.57$ \down{14}\\
  & Real $+$ SAADi    & \best{$0.63$}  & \best{$0.90$} & \best{$0.76$} & \best{$0.61$} & \best{$0.76$} & $0.74$ \up{3} \\
\midrule
\multirow{3}{*}{Segformer} 
  & Only Real       & $0.64$ & \best{$0.92$}  & $0.76$ & \best{$0.65$} & $0.77$ & $0.75$ \\
  & Real $+$ SSI   & $0.65$ & $0.87$  & $0.62$ & $0.62$ & $0.78$ & $0.71$ \down{4} \\
  & Real $+$ SAADi   & \best{$0.67$}  & $0.91$ & \best{$0.77$} & $0.64$ & \best{$0.80$} & $0.76$ \up{1}\\
\midrule
\multirow{3}{*}{UPerNet} 
  & Only Real       & $0.60$ & $0.91$  & $0.74$ & $0.62$ & $0.71$ & $0.72$\\
  & Real $+$ SSI    & \best{$0.63$} & $0.58$ & $0.54$ & $0.55$ & $0.74$ & $0.61$ \down{11} \\
  & Real $+$ SAADi   & $0.61$ & \best{$0.93$}  & \best{$0.76$} & \best{$0.63$} & \best{$0.76$} & $0.74$ \up{2} \\
\bottomrule
\end{tabular}
}}
\end{center}
\end{table}

\begin{figure}[!htbp]
  \centering
  \includegraphics[height=6cm,width=0.90\textwidth]{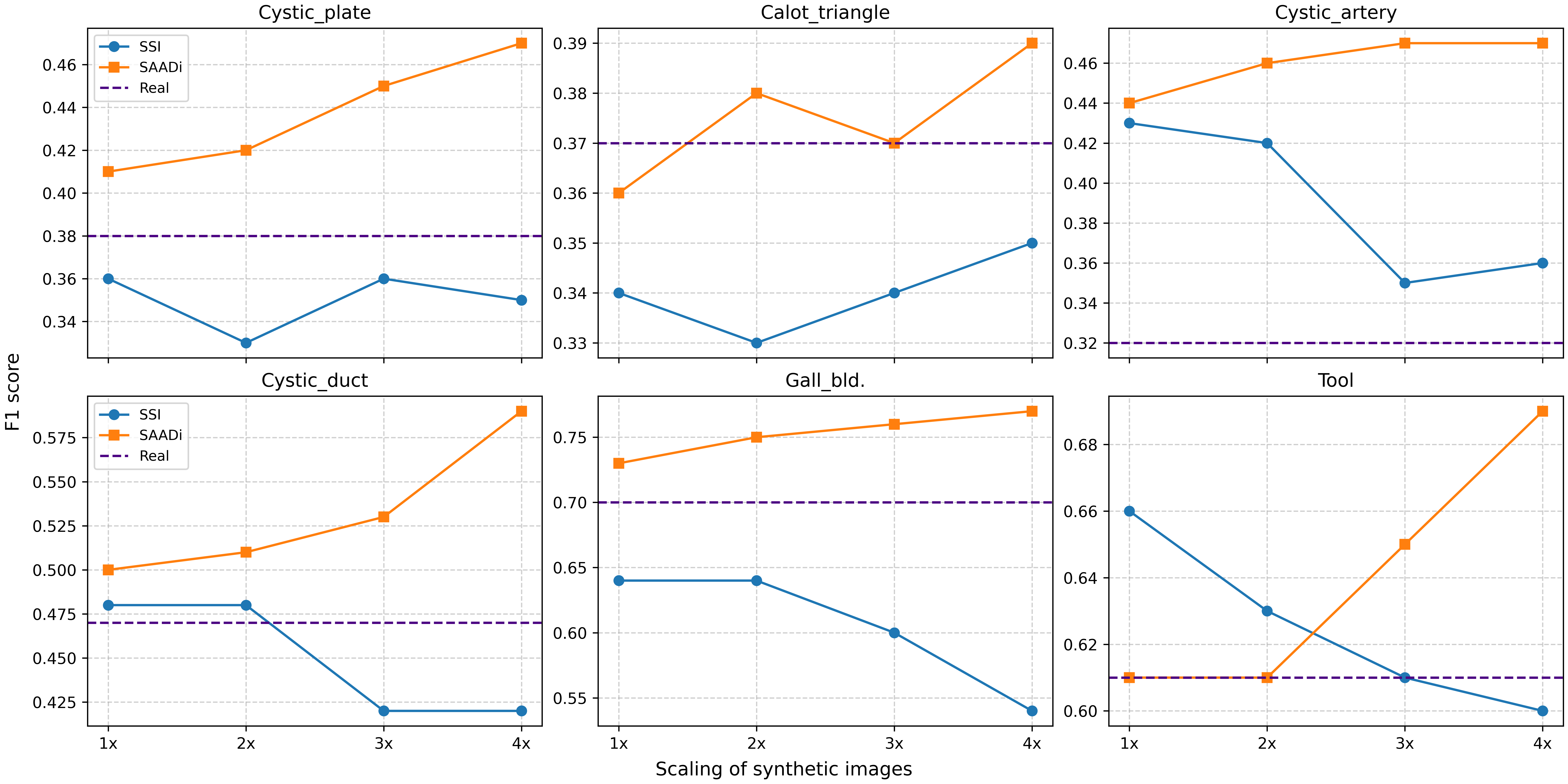}
  \caption{\textbf{Scaling of synthetic samples} for anatomy segmentation in the Endoscapes dataset. Adding synthetic samples from SAADi results in continuous performance improvements across classes. In contrast, samples from the SSI baseline lead to a decline in dice scores, reflecting inconsistencies from application-agnostic image generation.}
  \label{fig:data_scaling}
\end{figure}

In contrast, synthetic samples from SAADi achieve the best performance in five out of six classes, with gains ranging from $2$--$10\%$. Similar trends are observed on the AL dataset, where SAADi provides consistent improvements while SSI degrades performance. These results highlight an important observation: simply adding the same number of synthetic samples as the training set may not always yield large performance gains. Instead, aligning synthetic data with the downstream task, as in SAADi, is crucial for maximizing its utility. \textit{This further confirms that application-aware alignment is more critical for the surgical domain.}
\paragraph{Data scaling behavior} 
The scaling behavior of synthetic samples on the Ed dataset with the DV3 model is shown in Fig.~\ref{fig:data_scaling}, with additional results provided in the suppl. material. A consistent trend is that synthetic samples from SAADi yield steady improvements in performance across most classes. In contrast, samples from the SSI baseline lead to performance degradation beyond a specific scale for four classes, consistent with the observations of~\citet{azizi2023synthetic}. Another key observation is the plateauing of performance when more than $3\times$ or $4\times$ synthetic data is added. This saturation effect can be attributed to the fact that the diversity within the training distribution has already been extensively captured, and the generated samples largely reflect this existing diversity. To further improve performance, future work could explore incorporating variations in shape and texture depending on the inductive biases of the downstream model.
\textit{Our findings highlight the need for both task-alignment and data diversity in synthetic data for surgical applications.}
\paragraph{Iterative refinement of synthetic data} Tab.~\ref{tab:iter} indicates the results of a second round of refinement of synthetic samples generated by our approach, SAADi, on the Ed dataset. This additional step yields average performance improvements of $4$--$10\%$ across the different downstream models. The gains are modest for surgical tools, while the largest improvements are observed for the cystic plate and cystic duct classes. These findings suggest that refinement can be model-dependent, as each downstream architecture may exhibit its own inductive biases. Further exploration of multi-stage refinement could provide deeper insights into the limitations of aligning synthetic data generation with downstream tasks, which we leave for future work. Additional results are provided in the suppl. material. \textit{Overall, these results highlight iterative refinement as a promising strategy for enhancing the effectiveness of application-aligned synthetic data.}
\paragraph{Limitations} Although our approach is capable of generating synthetic images that benefit downstream tasks, certain limitations remain. First, SAADi requires a base diffusion model to generate the initial cohort of synthetic data. As a result, any biases present in the base model are propagated during fine-tuning and cannot be eliminated. Second, our approach relies on a selection model for classifying or segmenting synthetic images, which in turn requires annotated data. While self-supervised models may help alleviate this dependency, further investigation is needed. Third, although we employ lightweight fine-tuning, this step still adds to the computational cost of generation, which may hinder real-time applications. Future work could explore integrating feedback-guided approach~\citep{askari2025improving} with application alignment to reduce these overheads and further improve real-time applicability.

\begin{table}
\caption{\textbf{Iterative refinement of SAADi samples.} Results are reported on the Endoscapes dataset. $1^{st}$ denotes the first round of SAADi fine-tuning, and $2^{nd}$ indicates the second round with iterative refinement of synthetic samples. Refinement consistently improves performance across different models, with gains in the range of $4$--$10\%$.}
\label{tab:iter}
\begin{center}
{\small
\resizebox{\linewidth}{!}{
\begin{tabular}{l cc ccccccc}
\toprule
Method & Training data & Cystic plate & Calot triangle & Cystic artery &Cystic duct& Gall bladder& Tool & Mean\\
\midrule
\multirow{3}{*}{DV3} 
  & Only Real       & $0.38$ & \best{$0.37$} & $0.42$ & $0.47$ & $0.70$ & $0.61$ & $0.49$ \\
  & Real $+$ SAADi ($1^{st}$)   & $0.41$ & $0.36$ & $0.44$ & $0.50$ & \best{$0.73$} & $0.61$ & $0.51$ \up{2}\\
  & Real $+$ SAADi ($2^{nd}$)   & \best{$0.42$} & $0.36$ & \best{$0.47$} & \best{$0.53$} & \best{$0.73$} & \best{$0.64$} & $0.53$ \up{4} \\
\midrule
\multirow{3}{*}{Segformer} 
  & Only Real       & $0.40$ & $0.36$ & $0.30$ & $0.41$ & $0.68$ & $0.62$ & $0.46$ \\
  & Real $+$ SAADi ($1^{st}$)  & $0.41$ & $0.49$ & $0.40$ & \best{$0.52$} & $0.70$ & $0.63$ & $0.52$ \up{6} \\
  & Real $+$ SAADi ($2^{nd}$)   & \best{$0.43$} & \best{$0.52$} & \best{$0.41$} & \best{$0.52$} & \best{$0.73$} & \best{$0.64$} & $0.54$ \up{8}\\
\midrule
\multirow{3}{*}{Upernet} 
  & Only Real       & $0.33$ & $0.27$ & $0.37$ & $0.38$ & $0.54$ & $0.41$ & $0.38$\\
  & Real $+$ SAADi ($1^{st}$)   & $0.40$ & $0.44$ & $0.46$ & $0.40$ & $0.57$ & $0.61$ & $0.48$ \up{4} \\
  & Real $+$ SAADi ($2^{nd}$)   & \best{$0.41$} & \best{$0.45$} & \best{$0.48$} & \best{$0.46$} & \best{$0.61$} & \best{$0.64$} & $0.51$ \up{13} \\
\bottomrule
\end{tabular}
}}
\end{center}
\end{table}

\section{Conclusion}
In this work, we presented \textbf{SAADi}, an application-aligned diffusion framework for surgical image synthesis that explicitly adapts generation to downstream tasks. Instead of relying on human feedback, SAADi leverages downstream model evaluations to fine-tune diffusion models on pairs of preferred and non-preferred samples, producing synthetic data that is both realistic and task-relevant. Comprehensive experiments on three surgical datasets demonstrate consistent improvements in both classification and segmentation tasks, with notable gains for underrepresented classes. Furthermore, iterative refinement yields additional improvements, highlighting the importance of alignment beyond simple dataset scaling. Taken together, these results establish preference alignment as a promising direction for generating clinically indicative synthetic data and mitigating the challenge of data scarcity in surgical data science.

\paragraph{Acknowledgement} This work is partly supported by BMFTR (Federal Ministry of Research,
Technology and Space) in DAAD project 57616814 (SECAI, School of
Embedded Composite AI, https://secai.org/) as part of the program Konrad
Zuse Schools of Excellence in Artificial Intelligence. Also partially funded by the German Research Foundation (DFG, Deutsche Forschungsgemeinschaft) as part of Germany’s Excellence Strategy – EXC 2050/1 –Project ID 390696704 – Cluster of Excellence “Centre for Tactile Internet with Human-in-the-Loop” (CeTI) of Technische Universit\"at Dresden.

\bibliographystyle{abbrvnat}
\bibliography{ref}

\clearpage
\appendix

\section{Supplementary Material}
\begin{figure}[H]
  \centering
  \includegraphics[height=5.5cm,width=0.90\textwidth]{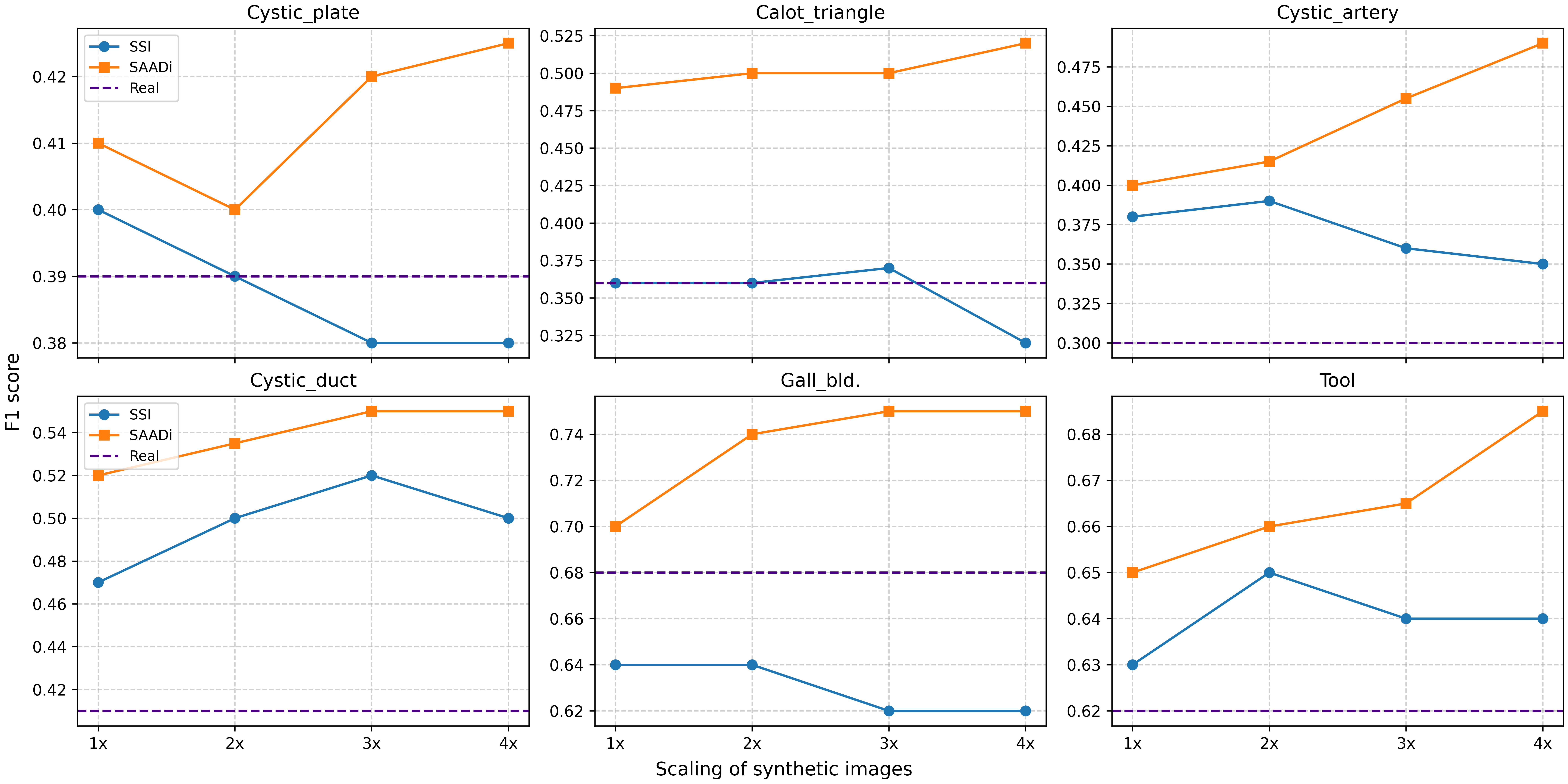}
  \caption{Scaling of synthetic samples for anatomy segmentation in the Endoscapes dataset with Segformer model. The addition of synthetic samples from SSI model leads to performance improvement in four out of six classes. The synthetic samples from SAADi leads to continuous gains in dice scores across different classes.}
  \label{fig:data_scaling_seg}
\end{figure}

\begin{figure}[H]
  \centering
  \includegraphics[height=4.5cm,width=0.95\textwidth]{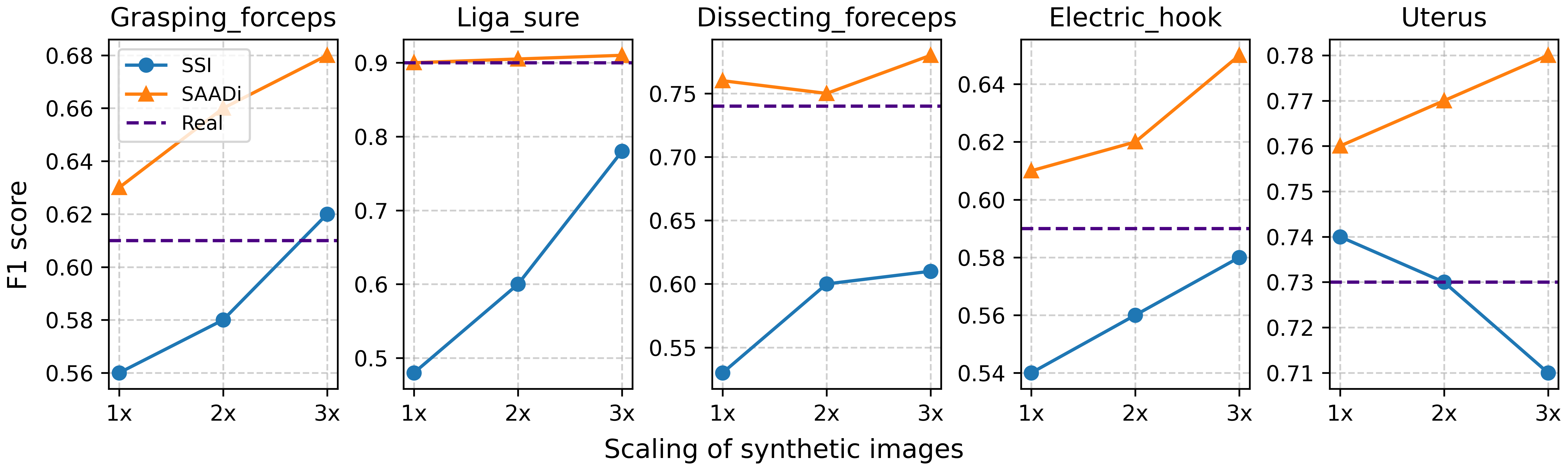}
  \caption{Scaling of synthetic samples for tool and anatomy segmentation on the Autolaparo dataset with DV$3$ model. For this dataset, in constract to Fig.~\ref{fig:data_scaling} and Fig.~\ref{fig:data_scaling_seg} the synthetic samples from the baseline model leads to continuous increase of F$1$ scores for all the tools. However, the samples from our appraoch SAADi are more aligned to task and hence they outperform the baseline and shows the best scores for all the classes. For the Liga sure instrument we notice the scores to plateua beyond $1\times$. }
\end{figure}

\begin{figure}[H]
  \centering
  \includegraphics[height=4.5cm,width=0.95\textwidth]{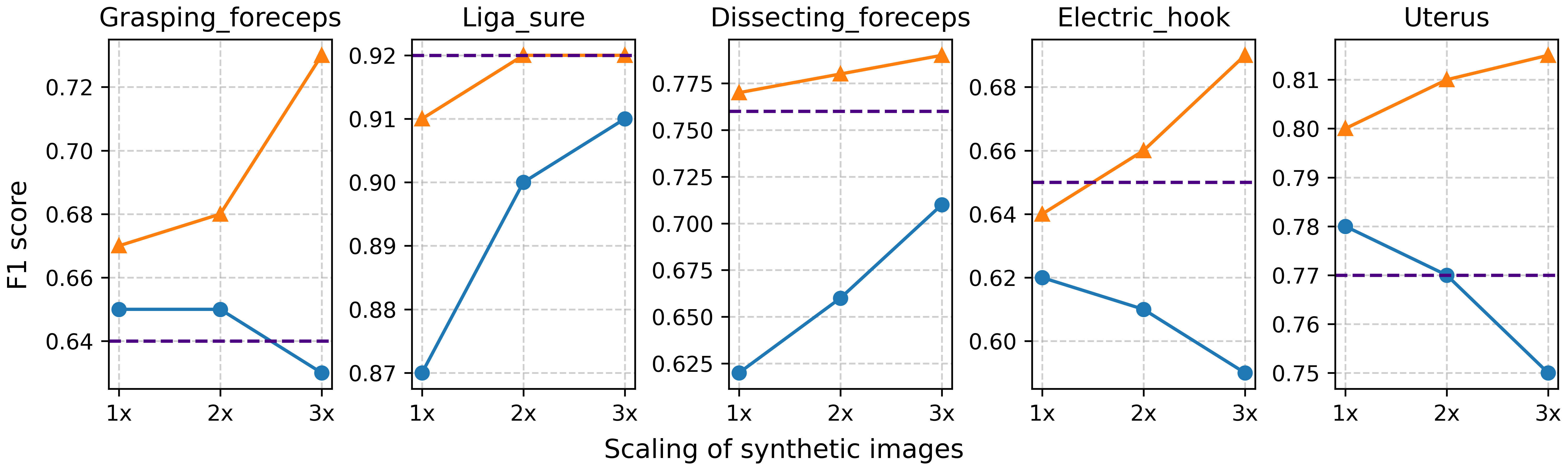}
  \caption{Scaling of synthetic samples for tool and organ segmentation on the Autolaparo dataset.}
\end{figure}


\end{document}